\documentclass[smallcondensed,natbib]{svjour3}
\usepackage{amssymb}
\usepackage{amsmath}
\usepackage{graphicx}
\usepackage{float}
\usepackage{geometry}
\usepackage{setspace}
\usepackage{bbm}
\usepackage[T1]{fontenc}
\usepackage[utf8]{inputenc}
\usepackage{lipsum}
\usepackage{caption}
\usepackage{bm}
\usepackage{algorithm}
\usepackage{algpseudocode}
\usepackage{url}
\usepackage{subfig}
\usepackage{array}
\usepackage{multirow}

\newcommand{\argmin}{\operatornamewithlimits{argmin}}
\renewcommand{\vec}[1]{\mathbf{#1}}

\allowdisplaybreaks

\begin{document}
\newcolumntype{C}{>{\centering\arraybackslash}p{5em}}

\title{Nearest Labelset Using Double Distances for Multi-label Classification}

\author{Hyukjun Gweon   \and
        Matthias Schonlau \and
        Stefan Steiner}

\institute{
            Hyukjun Gweon \at
              University of Waterloo, 200 University Avenue West, Waterloo, ON, Canada\\
              \email{hgweon@uwaterloo.ca}           
           \and
            Matthias Schonlau \at
              University of Waterloo, 200 University Avenue West, Waterloo, ON, Canada\\
              \email{schonlau@uwaterloo.ca}          
           \and
            Stefan Steiner \at
              University of Waterloo, 200 University Avenue West, Waterloo, ON, Canada\\
              \email{shsteiner@uwaterloo.ca}
}

\maketitle

\begin{abstract}
Multi-label classification is a type of supervised learning where an instance may belong to multiple labels simultaneously.
Predicting each label independently has been criticized for not exploiting any correlation between labels.
In this paper we propose a novel approach, Nearest Labelset using Double Distances ($NLDD$), that predicts the labelset observed in the training data that minimizes a weighted sum of the distances in both the feature space and the label space to the new instance.
The weights specify the relative tradeoff between the two distances. The weights are estimated from a binomial regression of the number of misclassified labels as a function of the two distances.
Model parameters are estimated by maximum likelihood.
$NLDD$ only considers labelsets observed in the training data, thus implicitly taking into account label dependencies.
Experiments on benchmark multi-label data sets show that the proposed method on average outperforms other well-known approaches in terms of $Hamming \ loss$, $0/1 \ loss$, and $multi$-$label$ $accuracy$ and ranks second after $ECC$ on the $F$-$measure$.
\keywords{Multi-label classification, Machine learning, Label correlations}
\end{abstract}

\section{Introduction}

In multi-label classification, an instance can belong to multiple labels at the same time. This is different from multi-class or binary classification, where an instance can only be associated with a single label. For example, a newspaper article talking about electronic books may be labelled with multiple topics such as business, arts and technology simultaneously. Multi-label classification has been applied in many areas of application including text \citep{Schapire2000, Godbole2004}, image \citep{Boutell2004, Zhang2007}, music \citep{Li2003, Trohidis2008} and bioinformatics \citep{Elisseeff2001}. A labelset for an instance is the set of all labels that are associated with that instance.

Approaches for solving multi-label classification problems may be categorized into either problem transformation methods or algorithm adaptation methods \citep{Tsoumakas2007a}. Problem transformation methods transform a multi-label problem into one or more single-label problems. For the single-label classification problems, binary or multi-class classifiers are used. The results are combined and transformed back into a multi-label representation. Algorithm adaptation methods, on the other hand, modify specific learning algorithms directly for multi-label problems. Individual approaches are explained in Section \ref{related}.

In this paper, we propose a new problem transformation approach to multi-label classification.
Our proposed approach applies the nearest neighbor method to predict the label with the shortest distance in the feature space.
However, because we have multiple labels, we additionally consider the shortest distance in the label space. We then find the labelset that minimizes the expected label misclassification rate as a function of both distances, feature space and label space, exploiting high-order interdependencies between labels. The nonlinear function is estimated using maximum likelihood.

The effectiveness of the proposed approach is evaluated with various multi-label data sets. Our experiments show that the proposed method performs on average better on standard evaluation metrics  ($Hammming \ loss$, $0/1 \ loss$, $multi$-$label$ $accuracy$ and the $F$-$measure$) than other commonly used algorithms.

The rest of this paper is organized as follows: In Section \ref{related} we review previous work on multi-label classification. In Section \ref{method}, we present the details of the proposed method. In Section \ref{experiments}, we report on experiments that compare the proposed method with other algorithms on standard metrics. In Section \ref{discussion} we discuss the results. In Section \ref{conclusion}, we draw conclusions.

\section{Related work} \label{related}

There are several approaches to classifying multi-label data. The most common approach, binary relevance ($BR$) \citep{Zhang2005, Tsoumakas2007a}, transforms a multi-label problem into separate binary problems. That is, using training data, $BR$ constructs a binary classifier for each label independently. For a test instance, the prediction set of labels is obtained simply by combining the individual binary results. In other words, the predicted labelset is the union of the results predicted from the $L$ binary models. This approach requires one binary model for each label. The method has been adapted in many domains including text \citep{Goncalves2003}, music \citep{Li2003} and images \citep{Boutell2004}. One drawback of the basic binary approach is that it does not account for any correlation that may exist between labels, because the labels are modelled independently. Taking correlations into account is often critical for prediction in multi-label problems \citep{Godbole2004, Ji2008}.

A method related to $BR$ is Subset-Mapping ($SMBR$) \citep{Schapire1999, Read2011}.
For a new instance, a vector of labels is obtained by the binary outputs of $BR$ and the final prediction is made by the training labelset with the shortest Hamming distance to the prediction set.
For predictions $SMBR$ only chooses labelsets observed in training data, thus $SMBR$ exploits the interdependencies among labels.

An extension of binary relevance is Classifier Chain ($CC$) \citep{Read2011}.
$CC$ fits labels sequentially using binary classifiers.
Labels already predicted are included as features in subsequent classifiers until all labels have been fit.
Including previous predictions as features ``chains'' the classifiers together and also takes into account potential label correlations.
However, the order of the labels in a chain affects the predictive performances.
\citet{Read2011} also introduced the ensemble of classifier chains ($ECC$), where multiple $CC$ are built with re-sampled training sets.
The order of the labels in each $CC$ is randomly chosen. The prediction label of an $ECC$ is obtained by the majority vote of the $CC$ models.

Label Powerset learning ($LP$) transforms a multi-label classification into a multi-class problem \citep{Tsoumakas2007a}. In other words, $LP$ treats each labelset as a single label. The transformed problem requires a single classifier. Although $LP$ captures correlations between labels, the number of classes in the transformed problem increases exponentially with the number of original labels. $LP$ learning can only choose observed labelsets for predictions \citep{Tsoumakas2007a, Read2008}.

The random k-labelsets method, ($RAKEL$) \citep{Tsoumakas2007b}, is a variation on the $LP$ approach.
In a multi-label problem with $L$ different labels, $RAKEL$ employs $m$ multi-class models each of which considers $k (\le L)$ randomly chosen labels, rather than the entire labelset. For a test instance, the prediction labelset is obtained by the majority vote of the results based on the $m$ models. $RAKEL$ overcomes the problem that the number of multinomial classes increases exponentially as a function of the number of labels. It also considers interdependencies between labels by using multi-class models with subsets of the labels.

A popular lazy learning algorithm based on the $k$ Nearest Neighbours ($kNN$) approach is $MLKNN$ \citep{Zhang2007}. Like other $kNN$-based methods, $MLKNN$ identifies the $k$ nearest training instances in the feature space for a test instance. Then for each label, $MLKNN$ estimates the prior and likelihood for the number of neighbours associated with the label. Using Bayes theorem, $MLKNN$ calculates the posterior probability from which a prediction is made.

The Conditional Bernoulli Mixtures ($CBM$) \citep{li2016} approach transforms a multi-label problem into a mixture of binary and multi-class problems. $CBM$ divides the feature space into $K$ regions and learns a multi-class classifier for the regional components as well as binary classifiers in each region. The posterior probability for a labelset is obtained by mixing the multi-class and multiple binary classifiers. The model parameters are estimated using the Expectation Maximization algorithm.

\section{The nearest labelset using double distances approach} \label{method}

\subsection{Hypercube view of a multi-label problem}

In multi-label classification, we are given a set of possible output labels $\mathcal{L} = \{1, 2, ..., L\}$.
Each instance with a feature vector $\vec{x} \in \mathbb{R}^{d}$ is associated with a subset of these labels.
Equivalently, the subset can be described as $\vec{y} = (y^{(1)},y^{(2)},...,y^{(L)})$, where $y^{(i)} = 1$ if label $i$ is associated with the instance and $y^{(i)} = 0$ otherwise.
A multi-label training data set is described as $T = \{(\vec{x}_{i},\vec{y}_{i}), i=1, 2, ...,N\}$.

Any labelset $\vec{y}$ can be described as a vertex in the $L$-dimensional unit hypercube \citep{Tai2012}. Each component $y^{(i)}$ of $\vec{y}$ represents an axis of the hypercube. As an example, Figure 1 illustrates the label space of a multi-label problem with three labels ($y^{(1)}$, $y^{(2)}$, $y^{(3)}$).

Assume that the presence or absence of each label is modeled independently with a probabilistic classifier. For a new instance, the classifiers provide the probabilities, $p^{(1)}$, ..., $p^{(L)}$, that the corresponding labels are associated with the instance. Using the probability outputs, we may obtain a $L$-dimensional vector $\hat{\vec{p}} = (p^{(1)},p^{(2)},...,p^{(L)})$.
Every element of $\hat{\vec{p}}$ has a value from 0 to 1 and the vector $\hat{\vec{p}}$ is an inner point in the hypercube (see Figure 1). Given $\hat{\vec{p}}$ the prediction task is completed by assigning the inner point to a vertex of the cube.

For the new instance, we may calculate the Euclidean distance, $D_{\vec{y}_{i}}$, between $\hat{\vec{p}}$ and each $\vec{y}_{i}$ (i.e. the labelset of the $i^{th}$ training instance). In Figure 1, three training instances $\vec{y}_{1}$, $\vec{y}_{2}$ and $\vec{y}_{3}$ and the corresponding distances are shown. A small distance $D_{\vec{y}_{i}}$ indicates that $\vec{y}_{i}$ is likely to be the labelset for the new instance.

\begin{figure}[h]
\centering
  \includegraphics[scale=0.5]{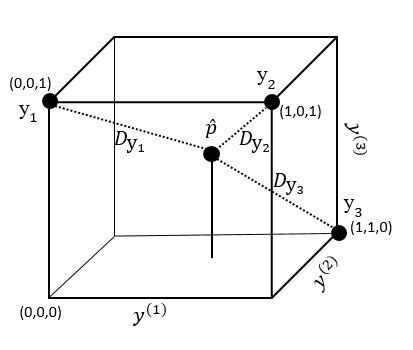}
  \caption{An illustration of the label space when $L=3$. Each vertex represents a labelset. The inner point represents a fitted vector of an instance. $D_{\vec{y}_{i}}$ represents the distance between $\hat{\vec{p}}$ and $\vec{y}_{i}$.}
\end{figure}

\subsection{Nearest labelset using double distances ($NLDD$)} \label{NLDD}

In addition to computing the distance in the label space, $D_{\vec{y}_{i}}$, we may also obtain the (Euclidean) distance in the feature space, denoted by $D_{\vec{x}_{i}}$.
The proposed method, $NLDD$, uses both $D_{\vec{x}}$ and $D_{\vec{y}}$ as predictors to find a training labelset that minimizes the expected loss.
For each test instance, we define loss as the number of misclassified labels out of $L$ labels.
The expected loss is then $L \theta$ where $\theta = g(D_{\vec{x}},D_{\vec{y}})$ represents the probability of misclassifying each label.
The predicted labelset, $\hat{\vec{y}}^{*}$, is the labelset observed in the training data that minimizes the expected  loss:
\begin{align}
\hat{\vec{y}}^{*} = \underset{\vec{y} \in T}\argmin \ L \hskip 0.05cm g(D_{\vec{x}},D_{\vec{y}})
\label{eq:min}
\end{align}
The loss follows a binomial distribution with $L$ and a parameter $\theta$.
We model $\theta = g(D_{\vec{x}},D_{\vec{y}})$ as follows:

\begin{align}
\log\left(\frac{\theta}{1-\theta}\right) = \beta_{0} + {\beta_{1}} D_{\vec{x}} + {\beta_{2}} D_{\vec{y}}
\label{eq:binomial}
\end{align}
where $\beta_{0}$, $\beta_{1}$ and $\beta_{2}$ are the model parameters.
Greater values for $\beta_{1}$ and $\beta_{2}$ imply that $\theta$ becomes more sensitive to the distances in the feature and label spaces, respectively.
The misclassification probability decreases as $D_{\vec{x}}$ and $D_{\vec{y}}$ approach zero.

A test instance with $D_{\vec{x}} = D_{\vec{y}} = 0$  has a duplicate instance in the training data (i.e. with identical features).
The predicted  probabilities for the test instance are either 0 or 1 and the match the labels of the duplicate training observation.
For such a ``double''-duplicate instance (i.e. $D_{\vec{x}} = D_{\vec{y}} = 0$), the probability of misclassification is $ 1/(1+ e^{-\beta_{0}}) >0$.
As expected, the uncertainty of a test observation with a ``double-duplicate'' training observation is greater than zero.

The model in \eqref{eq:binomial} implies $g(D_{\vec{x}},D_{\vec{y}}) = 1/(1+ e^{-(\beta_{0} + {\beta_{1}} D_{\vec{x}} + {\beta_{2}} D_{\vec{y}})})$.
Because $\log\left(\frac{\theta}{1-\theta}\right)$ is a monotone transformation of $\theta$ and  $L$ is a constant, the minimization problem in \eqref{eq:min} is equivalent to
\begin{align}
\hat{\vec{y}}^{*} = \underset{\vec{y} \in T}\argmin \ {\beta_{1}} D_{\vec{x}} + {\beta_{2}} D_{\vec{y}}
\label{eq:weight}
\end{align}
That is, $NLDD$ predicts by choosing the labelset of the training instance that minimizes the weighted sum of the distances.
For prediction, the only remaining issue is how to estimate the weights.

\subsection{Estimating the relative weights of the two distances}
\begin{figure}[h]
\centering
  \includegraphics[scale=0.5]{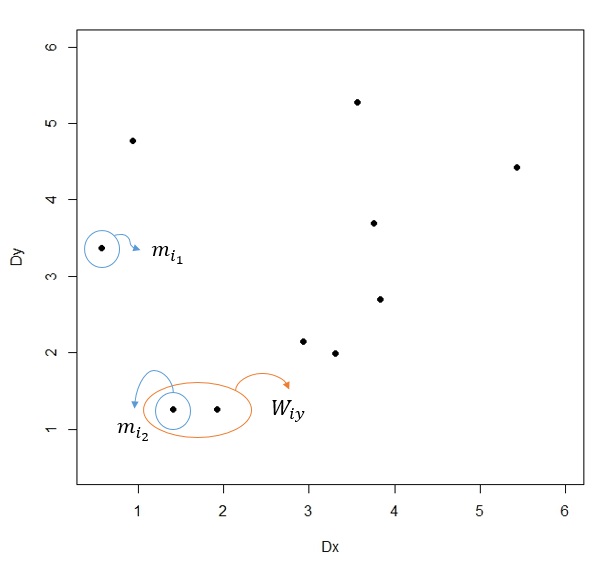}
  \caption{An illustration of how to identify $m_{i_{1}}$ and $m_{i_{2}}$ for $N=20$. $T_{1}$ and $T_{2}$ contain 10 instances each. The 10 points in the scatter plot were obtained by calculating $D_{x}$ and $D_{y}$ between an instance in $T_{2}$ and the 10 instances in $T_{1}$. In this example two points have the lowest distance in $D_{y}$ and are candidates for $m_{i_{2}}$. Among the candidates, the point with the lowest $D_x$ is chosen.}
\label{ex}
\end{figure}

We need to estimate the parameters $\beta_{0}, \beta_{1}$ and $\beta_{2}$.
This requires computing $D_{\vec{y}}$, but of course the outcomes in the test data are not known.
We therefore split the training data, $T$,
equally into two data sets, $T_{1}$ and $T_{2}$. $T_{2}$ is used for validation.
Using $T_{1}$, we next fit a binary classifier to each of the $L$ labels separately and obtain the labelset predictions (i.e. probability outcomes) for the instances in $T_{2}$.
We then create a set of $(D_{\vec{x}},D_{\vec{y}})$ by pairing instances in $T_{1}$ with those in $T_{2}$.
Note that matching any single instance in $T_{2}$ to those in $T_{1}$ results in $N/2$ distance pairs.
Most of the pairs are uninformative because the distance in either the feature space or the label space is very large.
Moreover, since $T_{2}$ contains $N/2$ instances, the number of possible pairs is potentially large ($N^2$/4).
Therefore, to reduce computational complexity, for each instance we only identify two pairs: the pair with the smallest distance in $\vec{x}$ and the pair with the smallest distance in $\vec{y}$.
In case of ties in one distance, the pair with the smallest value in the other distance is chosen. More formally we identify the first pair $m_{i_{1}}$ by
$$
m_{i_{1}} = \underset{(D_{x}, D_{y}) \in W_{ix}}\argmin D_{y}
$$
where $W_{ix}$ is the set of  pairs that are tied; i.e. that each corresponds to the minimum distance in $D_x$.
Similarly, the second pair $m_{i_{2}}$ is found by
$$
m_{i_{2}} = \underset{(D_{x}, D_{y}) \in W_{iy}}\argmin D_{x}.
$$
where $W_{iy}$ is the set of labels that  are tied with the minimal distance in $D_y$.
Figure \ref{ex} illustrates an example of how to identify $m_{i_{1}}$ and $m_{i_{2}}$ for $N=20$.
Our goal was to identify the instance with the smallest distance in $\vec{x}$ and the instance with the smallest distance in $\vec{y}$.
Note that $m_{i_{1}}$ and $m_{i_{2}}$ may be the same instance
If we find a single  instance that minimizes both distances, we use just that instance. (A possible duplication of that instance is unlikely to make any difference in practice).

The two pairs corresponding to the $i^{th}$ instance in $T_{2}$ are denoted as the set $S_{i} = \left\{m_{i_{1}}, m_{i_{2}} \right\}$, and their union for all instances is denoted as $S = \bigcup_{i=1}^{N/2} S_{i}$.
The binomial regression specified  in \eqref{eq:binomial}  is performed on the instances in $S$ and maximum likelihood estimators of the parameters are obtained.
Algorithm \ref{alg:train} outlines the training procedure.

\begin{algorithm}[h]
\begin{algorithmic}
\renewcommand{\thealgorithm}{}
\State{\textbf{Input:} training data $T$, number of labels $L$}
\State{\textbf{Output:} probabilistic classifiers $h^{(i)}$, binomial regression $g$}
\State{Split $T$ into $T_{1}$ and $T_{2}$}
\For{$i=1$ to $L$}
\State train probabilistic classifier $h^{(i)}$ based on $T$
\State train probabilistic classifier $h^{(i)}_{*}$ based on $T_{1}$
\EndFor
\State{$S,W \leftarrow \emptyset$}
\For{each instance in  $T_{2}$}
\State obtain $\hat{\vec{p}} = (h^{(1)}_{*}(\vec{x}),...,h^{(L)}_{*}(\vec{x}))$
\For{each instance in  $T_{1}$}
\State compute $D_{\vec{x}}$ and $D_{\vec{y}}$
\State $W \leftarrow W \cup (D_{\vec{x}},D_{\vec{y}})$
\EndFor
\State find $m_{1}, m_{2} \in W$
\State update $S \leftarrow S \cup \left\{m_{1},m_{2}\right\}$
\EndFor
\State Fit $\log\left(\frac{\theta}{1-\theta}\right) = \beta_{0} + {\beta_{1}}D_{\vec{x}} + {\beta_{2}}D_{\vec{y}}$ to $S$
\State Obtain $g: S \rightarrow \hat{\theta} = \frac{e^{\hat{f}}}{1+e^{\hat{f}}}$ where
$\hat{f} = \hat{\beta}_{0} + {\hat{\beta}_{1}} D_{\vec{x}} + {\hat{\beta}_{2}} D_{\vec{y}}$
\end{algorithmic}
\caption{The training process of $NLDD$}
\label{alg:train}
\end{algorithm}

For the classification of a new instance, we first obtain $\hat{\vec{p}}$ using the probabilistic classifiers fitted to the training data $T$.
$D_{\vec{x}_{j}}$ and $D_{\vec{y}_{j}}$ are obtained by matching the instance with the $j^{th}$ training instance.
Using the $MLEs$ $\hat{\beta}_{0}$, $\hat{\beta}_{1}$ and $\hat{\beta}_{2}$, we calculate
$\hat{\theta}_{j} = \frac{e^{\hat{f}_{j}}}{1+e^{\hat{f}_{j}}}$ where $\hat{f}_{j} = \hat{\beta}_{0} + {\hat{\beta}_{1}}D_{\vec{x}_{j}} + {\hat{\beta}_{2}}D_{\vec{y}_{j}}$.
The final prediction of the new instance is obtained by
$$
\hat{\vec{y}} = \underset{\vec{y}_{j} \in T}\argmin \ \hat{E}(loss) = \underset{\vec{y}_{j} \in T}\argmin \ \hat{\theta}_{j}.
$$
The second equality holds because $\hat{E}(loss) = L\hat{\theta}$ and $L$ is a constant.
As in $LP$, $NLDD$ chooses a training labelset as the predicted vector.
Algorithm \ref{alg:classify} outlines the classification procedure.

\begin{algorithm}[h]
\begin{algorithmic}
\State{\textbf{Input:} new instance $\vec{x}$, binomial model $g$, probabilistic classifiers $h^{(i)}$, training data $T$ of size $N$}
\State{\textbf{Output:} multi-label classification vector $\hat{\vec{y}}$}
\For{$j=1$ to $N$}
\State compute $\hat{\vec{p}} = (h^{(1)}(\vec{x}),...,h^{(L)}(\vec{x}))$
\State compute $D_{\vec{x}_{j}}$ and $D_{\vec{y}_{j}}$
\State obtain $\hat{\theta}_{j} \leftarrow g(D_{\vec{x}_{j}},D_{\vec{y}_{j}})$
\EndFor
\State return $\hat{\vec{y}} \leftarrow \underset{\vec{y}_{j} \in T}\argmin \ \hat{\theta}_{j}$
\end{algorithmic}
\caption{The classification process of $NLDD$}
\label{alg:classify}
\end{algorithm}

The training time of $NLDD$ is $O(L  (f(d,N) + f(d,N/2) + g(d,N/2)) + N^{2}(d+L) + N log(k))$ where $O(f(d,N))$ is the complexity of each binary classifier with $d$ features and $N$ training instances, $O(g(d,N/2))$ is the complexity for predicting each label for $T_{2}$, $N^{2}(d+L)$ is the complexity for obtaining the distance pairs for the regression and $O(N log(k))$ is the complexity for fitting a binomial regression.
$T_{1}$ and $T_{2}$ have $N/2$ instances respectively.
$O(L f(d,N/2))$ is the complexity for fitting binary classifiers using $T_{1}$ and obtaining the probability results for $T_{2}$ takes $O(L g(d,N/2))$.
For each instance of $T_{2}$, we obtain $N/2$ numbers of distance pairs. This has complexity $O((N/2) (d+L))$. Since there are $N/2$ instances, overall it takes $O((N/2) (N/2) (d + L))$ or $O(N^2 (d+L))$ when omitting the constant.
Among the $N/2$ pairs for each instance of $T_{2}$, we only identify at most 2 pairs. This implies $N/2 \le s \le N$ where $s$ is the number of elements in $S$.
Each iteration of the Newton-Raphson method has a complexity of $O(N)$.
For $k$-digit precision complexity $O(log k)$ is required \citep{ypma1995}. Combined, the complexity for estimating the parameters with $k$-digit precision is   $O(N log(k))$.
In practice, however, this term is dominated by $N^{2}(d+L)$ as we can set $k << N$.

\section{Experimental evaluation} \label{experiments}

In this section we compare the algorithms for multi-label classification on nine data sets in terms of $Hamming \ loss$, $0/1 \ loss$, $multi$-$label$ $accuracy$ and $F$-$measure$.
We next introduce the data sets and the evaluation measures and then present the results of our experiments.

\subsection{Data sets} \label{dataset}

We evaluated the proposed approach using nine commonly used multi-label data sets from different domains. Table \ref{table:data} shows basic statistics for each data set including its domain, numbers of labels and features.
In the text data sets, all features are categorical (i.e. binary).
The last column ``lcard'', short for label cardinality, represents the average number of labels associated with an instance.
The data sets are ordered by ($|L| \cdot |X| \cdot |E|$).

The $emotions$ data set \citep{Trohidis2008} consists of pieces of music with rhythmic and timbre features. Each instance is associated with up to 6 emotion labels such as ``sad-lonely'', ``amazed-surprised'' and ``happy-pleased''.
The $scene$ data set \citep{Boutell2004} consists of images with 294 visual features. Each image is associated with up to 6 labels including ``mountain'', ``urban'' and ``beach''.
The $yeast$ data set \citep{Elisseeff2001} contains 2417 yeast genes in the Yeast Saccharomyces Cerevisiae. Each gene is represented by 103 features and is associated with a subset of 14 functional labels.
The $medical$ data set consists of documents that describe patient symptom histories. The data were made available in the Medical Natural language Processing Challenge in 2007. Each document is associated with a set of 45 disease codes.
The $slashdot$ data set consists of 3782 text instances with 22 labels obtained from Slashdot.org.
The $enron$ data set \citep{Klimt2004} contains 1702 email messages from the Enron corporation employees. The emails were categorized into 53 labels.
The $ohsumed$ data set \citep{hersh1994} is a collection of medical research articles from MEDLINE database. We used the same data set as in \citet{Read2011} that contains 13929 instances and 23 labels.
The $tmc2007$ data set \citep{srivastava2005} contains 28596 aviation safety reports associated with up to 22 labels.
Following \citet{Tsoumakas2011}, we used a reduced version of the data set with 500 features.
The $bibtex$ data set \citep{katakis08} consists of 7395 bibtex entries for automated tag suggestion. The entries were classified into 159 labels.
All data sets are available online at: http://mulan.sourceforge.net/datasets-mlc.html and http://meka.sourceforge.net/\#datasets.

\begin{table}[h]
\centering
\begin{tabular}{ |r|r|r|r|r|r| }
    \hline
    name &  domain & labels ($|L|$) & features ($|X|$) & examples ($|E|$) & lcards \\
    \hline
    emotions & music & 6 & 72 & 593 & 1.87 \\
    scene & image & 6 & 294 & 2407 & 1.07 \\
    yeast & biology & 14 & 103 & 2417 & 4.24 \\
    medical & text & 45 & 1449 & 978 & 1.25 \\
    slashdot & text & 22 & 1079 & 3782 & 1.18 \\
    enron & text & 53 & 1001 & 1702 & 3.37 \\
    ohsumed & text & 23 & 1002 & 13929 & 1.66 \\
    tmc2007 & text & 22 & 500 & 28596 & 2.16 \\
    bibtex & text & 159 & 1836 & 7395 & 2.40 \\
    \hline
\end{tabular}
\caption{Multi-label data sets and their associated characteristics. Label cardinality (lcards) is the average number of labels associated with an instance}
\label{table:data}
\end{table}

\subsection{Evaluation metrics}

Multi-label classifiers can be evaluated with various loss functions. Here, four of the most popular criteria are used: $Hamming \ loss$, $0/1 \ loss$, $multi$-$label$ $accuracy$ and $F$-$measure$.
These criteria are defined in the following paragraphs.

Let $L$ be the number of labels in a multi-label problem.
For a particular test instance, let $\vec{y} = (y^{(1)},...,y^{(L)})$ be the labelset where $y^{(j)}=1$ if the $j^{th}$ label is associated with the instance and 0 otherwise.
Let $\hat{\vec{y}} = (\hat{y}^{(1)},...,\hat{y}^{(L)})$ be the predicted values obtained by any machine learning method. $Hamming \ loss$ refers to the percentage of incorrect labels. The $Hamming \ loss$ for the instance is
$$
\text{$Hamming \ loss$} = 1 - \frac{1}{L} \sum^{L}_{j=1}  \mathbbm{1} \{ y^{(j)} = \hat{y}^{(j)} \}
$$
where $\mathbbm{1}$ is the indicator function.
Despite its simplicity, the $Hamming \ loss$ may be less discriminative than other metrics.
In practice, an instance is usually associated with a small subset of labels.
As the elements of the $L$-dimensional label vector are mostly zero, even the empty set (i.e. zero vector) prediction may lead to a decent $Hamming \ loss$.

The $0/1 \ loss$ is 0 if all predicted labels match the true labels and 1 otherwise. Hence,
$$
\text{$0/1 \ loss$} = 1 - \mathbbm{1}\{ \vec{y} = \hat{\vec{y}} \}.
$$
Compared to other evaluation metrics, $0/1$ $loss$ is strict as all the $L$ labels must match to the true ones simultaneously.

The $multi$-$label$ $accuracy$ \citep{Godbole2004} (also known as the $Jaccard$ index) is defined as the number of labels counted in the intersection of the predicted and true labelsets divided by the number of labels counted in the union of the labelsets. That is,
$$
\text{$Multi$-$label$ $accuracy$} = \frac{|\vec{y} \cap \hat{\vec{y}}|}{|\vec{y} \cup \hat{\vec{y}}|}.
$$
The $multi$-$label$ $accuracy$ measures the similarity between the true and predicted labelsets.

The $F$-$measure$ is the harmonic mean of precision and recall. The $F$-$measure$ is defined as
$$
\text{$F$-$measure$} =  \frac{2 |\vec{y} \cap \hat{\vec{y}}|}{|\vec{y}| + |\hat{\vec{y}}|}.
$$

The metrics above were defined for a single instance.
On each metric, the overall value for an entire test data set is obtained by averaging out the individual values.

\subsection{Experimental setup}

We compared our proposed method against $BR$, $SMBR$, $ECC$, $MLKNN$, $RAKEL$ and $CBM$.
To train multi-label classifiers, the parameters recommended by the authors were used.
In the case of $MLKNN$, we set the number of neighbors and the smoothing parameter to 10 and 1 respectively.
For $RAKEL$, we set the number of separate models to $2L$ and the size of each sub-labelset to $3$.
For $ECC$, the number of $CC$ models for each ensemble was set to $10$.
On the larger data sets ($ohsumed$, $tmc2007$ and $bibtex$), we fit $ECC$ using reduced training data sets (75\% of the instances and 50\% of the features) as suggested in \citet{Read2011}. On the same data sets, we ran $NLDD$ using 70\% of the training data to reduce redundancy in learning.

For $NLDD$, we used support vector machines ($SVM$) \citep{Vapnik2000} as the base classifier on unscaled variables with a linear kernel and tuning parameter $C=1$.
The $SVM$ scores were converted into probabilities using Platt's method \citep{Platt1999}. $SVM$ was also used as the base classifier for $BR$,
$SMBR$, $ECC$ and $RAKEL$.
The analysis was conducted in $R$ \citep{R2014} using the $e1071$ package \citep{Meyer2014} for $SVM$.
For the data sets with less than 5,000 instances 10-fold cross validations ($CV$) were performed. On the larger data sets, we used 75/25 train/test splits. For fitting binomial regression models, we divided the training data sets at random into two parts of equal sizes.

For implementing $CBM$ we used a Java program developed by the authors. The default settings (e.g. logistic regression and 10 iterations for the $EM$ algorithm) were used on non-large data sets. For the large data sets $tmc2007$ and $bibtex$, the number of iterations was set to 5 and random feature reduction was applied as suggested by the developers. On each data set we used train/test split available at their website (https://github.com/cheng-li/pyramid).

We applied the Wilcoxon signed-rank test \citep{wilcoxon45,demsar06} to compare the methods over multiple data sets because unlike the t--test it does not make a distributional assumption.
Also, the Wilcoxon test is more robust to outliers than the t--test \citep{demsar06}. Each test was one-sided at significance level 0.05. In multi-label classification, the Wilcoxon signed-ranks test was employed by \citet{Tsoumakas2011}.

In $NLDD$, when calculating distances in the feature spaces we used the standardized features so that no particular features dominated distances. For a numerical feature variable $x$, the standardized variable $z$ is obtained by $z = ({x-\bar{x}})/{\text{sd}(x)}$ where $\bar{x}$ and $\text{sd}(x)$ are the mean and standard deviation of $x$ in the training data.

\subsection{Results}

Tables \ref{tab:hamming} to \ref{tab:F} summarize the results in terms of $Hamming \ loss$, $0/1 \ loss$, $multi$-$label$ $accuracy$ and $F$-$measure$, respectively.
We also ranked the algorithms for each metric. The Wilcoxon test results report whether or not any two methods were significantly different in their rankings across  data sets. The results are shown at the bottom of each table.
$NLDD$ achieved highest average ranks on $Hamming \ loss$, $0/1 \ loss$ and $multi$-$label$ $accuracy$ while $ECC$ achieved the highest average rank on the $F$-$measure$ with $NLDD$ taking the second place (and the difference between $ECC$ and $NLDD$ was not statistically significant).
$RAKEL$ achieved the second highest average rank on $Hamming \ loss$, while $CBM$ achieved the second highest average rank on $0/1 \ loss$ and $multi$-$label$ $accuracy$.
The performance of $CBM$ on the $0/1 \ loss$ was very variable achieving the highest rank on five out of nine data sets and the second worst on two data sets.

Table \ref{tab:time2} shows the running time in seconds of the methods.
On the non-large data sets, the relative differences of running time between $NLDD$ and $BR$ tended to increase with the size of the data sets.
On two of the large data sets, $ohsumend$ and $tmc2007$, $NLDD$ required less time than $BR$ as we only used 70\% of the training data.

\begin{table}[ht]
\centering
\resizebox{\textwidth}{!}{
\begin{tabular}{|C|C|C|C|C|C|C|C|}
\hline
Data     & $BR$ & $SMBR$ & \boldmath{$NLDD$} & $ECC$ & $RAKEL$ & $MLKNN$ & $CBM$ \\ \hline
emotions & 0.1964(3) & 0.1995(4) & 0.1901(1) & 0.2010(5) & 0.1952(2) & 0.2646(6) & 0.3366(7) \\ \hline
scene    & 0.1042(6) & 0.1298(7) & 0.0948(4) & 0.0939(3) & 0.0895(1) & 0.0903(2) & 0.0953(5) \\ \hline
yeast    & 0.1990(4) & 0.2048(5) & 0.1902(1) & 0.2056(6) & 0.1964(3) & 0.1952(2) & 0.2130(7)  \\ \hline
medical  & 0.0096(3) & 0.0111(6) & 0.0097(4) & 0.0091(2) & 0.0097(5) & 0.0153(7) & 0.0086(1) \\ \hline
slashdot & 0.0467(4) & 0.0541(7) & 0.0452(3) & 0.0473(5) & 0.0439(2) & 0.0518(6) & 0.0436(1) \\ \hline
enron    & 0.0578(7) & 0.0563(6) & 0.0550(4) & 0.0528(2) & 0.0552(5) & 0.0526(1) & 0.0531(3) \\ \hline
ohsumed  & 0.0670(4) & 0.0717(6) & 0.0630(2) & 0.0737(7) & 0.0605(1) & 0.0697(5) & 0.0638(3) \\ \hline
tmc2007  & 0.0583(1) & 0.0587(2) & 0.0595(4) & 0.0633(5) & 0.0588(3) & 0.0706(7) & 0.0699(6) \\ \hline
bibtex   & 0.0158(7) & 0.0151(6) & 0.0134(1) & 0.0147(5) & 0.0150(4) & 0.0139(3) & 0.0138(2) \\ \hline
av. ranks & 4.3 & 5.4 & 2.7 & 4.4 & 2.9 & 4.3 & 3.9 \\ \hline\hline
\multicolumn{1}{|l|}{Significance} & \multicolumn{7}{l|}{\small\vtop{\hbox{\strut $NLDD > \{BR, SMBR, ECC, MLKNN\}; BR > \{SMBR\}$}\hbox{\strut  $RAKEL > \{BR, SMBR, ECC\}$}}} \\ \hline
\end{tabular}
}
\caption{$Hamming \ loss$ (lower is better) averaged over 10 cross validations (with ranks in parentheses).
The data sets are ordered as in Table \ref{table:data}. The results from the Wilcoxon test on whether or not any two results are  statistically  significant from one another  are summarized at the bottom of the table.}
\label{tab:hamming}
\end{table}

\begin{table}[ht]
\centering
\resizebox{\textwidth}{!}{
\begin{tabular}{|C|C|C|C|C|C|C|C|}
\hline
Data     & $BR$ & $SMBR$ & \boldmath{$NLDD$} & $ECC$ & $RAKEL$ & $MLKNN$ & $CBM$\\ \hline
emotions & 0.7181(5) & 0.7080(3) & 0.6900(2) & 0.7100(4) & 0.6793(1) & 0.8850(7) & 0.7980(6) \\ \hline
scene    & 0.4674(7) & 0.4242(6) & 0.3190(1) & 0.3511(3) & 0.3640(4) & 0.3702(5) & 0.3211(2) \\ \hline
yeast    & 0.8940(7) & 0.8180(6) & 0.7484(1) & 0.7977(3) & 0.8130(4) & 0.8179(5) & 0.7514(2) \\ \hline
medical  & 0.3191(6) & 0.3068(4) & 0.2792(2) & 0.3017(3) & 0.3191(5) & 0.4940(7) & 0.2263(1) \\ \hline
slashdot & 0.6452(6) & 0.6253(4) & 0.5232(2) & 0.6000(3) & 0.6277(5) & 0.9386(7) & 0.5127(1) \\ \hline
enron    & 0.9059(6) & 0.8765(3) & 0.8657(2) & 0.8788(4) & 0.9000(5) & 0.9588(7) & 0.8300(1) \\ \hline
ohsumed  & 0.7990(5) & 0.7872(4) & 0.7462(2) & 0.8193(6) & 0.7742(3) & 0.9495(7) & 0.7338(1) \\ \hline
tmc2007  & 0.7063(4) & 0.7043(3) & 0.7030(2) & 0.7316(5) & 0.7026(1) & 0.7732(7) & 0.7360(6) \\ \hline
bibtex   & 0.8504(6) & 0.8201(3) & 0.8081(2) & 0.8391(4) & 0.8413(5) & 0.9441(7) & 0.7815(1) \\ \hline
av. ranks & 5.8 & 4.0 & 1.8 & 3.9 & 3.7 & 6.6 & 2.3 \\ \hline\hline
\multicolumn{1}{|l|}{Significance} & \multicolumn{7}{l|}{\small\vtop{\hbox{\strut $NLDD > \{BR, SMBR, ECC, RAKEL, MLKNN\}; BR > \{MLKNN\}; SMBR > \{BR, MLKNN\};$}\hbox{\strut  $ECC > \{MLKNN\}; RAKEL > \{BR, MLKNN\}; CBM > \{BR, SMBR, MLKNN\}$}}} \\ \hline
\end{tabular}
}
\caption{$0/1 \ loss$ (lower is better) averaged over 10 cross validations (with ranks in parentheses). The loss is 0 if a predicted labelset matches the true labelset exactly and 1 otherwise. The results from the Wilcoxon test on whether or not any two results are  statistically  significant from one another  are summarized at the bottom of the table.}
\label{tab:zero}
\end{table}

\begin{table}[ht]
\centering
\resizebox{\textwidth}{!}{
\begin{tabular}{|C|C|C|C|C|C|C|C|}
\hline
Data     & $BR$ & $SMBR$ & \boldmath{$NLDD$} & $ECC$ & $RAKEL$ & $MLKNN$ & $CBM$\\ \hline
emotions & 0.5248(5) & 0.5467(4) & 0.5624(1) & 0.5587(2) & 0.5548(3) & 0.3253(7) & 0.4033(6) \\ \hline
scene    & 0.6357(7) & 0.6512(6) & 0.7422(1) & 0.6985(4) & 0.6990(3) & 0.6900(5) & 0.7178(2) \\ \hline
yeast    & 0.4992(7) & 0.5092(6) & 0.5461(1) & 0.5428(2) & 0.5194(4) & 0.5103(5) & 0.5216(3) \\ \hline
medical  & 0.7655(5) & 0.7696(4) & 0.7991(2) & 0.7934(3) & 0.7643(6) & 0.5787(7) & 0.8167(1) \\ \hline
slashdot & 0.4517(6) & 0.4687(4) & 0.5354(2) & 0.5067(3) & 0.4577(5) & 0.0694(7) & 0.5495(1) \\ \hline
enron    & 0.3974(6) & 0.4226(3) & 0.4122(4) & 0.4708(1) & 0.4088(5) & 0.3175(7) & 0.4297(2) \\ \hline
ohsumed  & 0.3848(6) & 0.3968(4) & 0.4105(3) & 0.4316(2) & 0.3940(5) & 0.0798(7) & 0.4918(1) \\ \hline
tmc2007  & 0.5750(3) & 0.5784(2) & 0.5692(4) & 0.5670(5) & 0.5710(1) & 0.4719(7) & 0.5186(6) \\ \hline
bibtex   & 0.3259(6) & 0.3387(3) & 0.3492(2) & 0.3321(4) & 0.3335(5) & 0.1281(7) & 0.3761(1) \\ \hline
av. ranks & 5.7 & 4.0 & 2.3 & 2.9 & 4.1 & 6.6 & 2.6 \\ \hline\hline
\multicolumn{1}{|l|}{Significance} & \multicolumn{7}{l|}{\small\vtop{\hbox{\strut $NLDD > \{BR, SMBR, RAKEL, MLKNN\}; BR > \{MLKNN\}; SMBR > \{BR, MLKNN\};$}\hbox{\strut  $ECC > \{BR, SMBR, RAKEL, MLKNN\}; RAKEL > \{BR, MLKNN\}; CBM > \{MLKNN\}$}}} \\ \hline
\end{tabular}
}
\caption{$Multi$-$label$ $accuracy$ (higher is better) averaged over 10 cross validations (with ranks in parentheses). The results from the Wilcoxon test on whether or not any two results are  statistically  significant from one another  are summarized at the bottom of the table.}
\label{tab:accuracy}
\end{table}

\begin{table}[ht]
\centering
\resizebox{\textwidth}{!}{
\begin{tabular}{|C|C|C|C|C|C|C|C|}
\hline
Data     & $BR$ & $SMBR$ & \boldmath{$NLDD$} & $ECC$ & $RAKEL$ & $MLKNN$ & $CBM$\\ \hline
emotions & 0.6033(5) & 0.6291(4) & 0.6446(2) & 0.6477(1) & 0.6316(3) & 0.3989(7) & 0.4723(6) \\ \hline
scene    & 0.6245(7) & 0.6429(6) & 0.7358(1) & 0.7152(3) & 0.6922(4) & 0.6833(5) & 0.7307(2) \\ \hline
yeast    & 0.6094(7) & 0.6159(4) & 0.6438(2) & 0.6465(1) & 0.6249(3) & 0.6140(6) & 0.6154(5) \\ \hline
medical  & 0.7945(5) & 0.7957(4) & 0.8268(2) & 0.8257(3) & 0.7928(6) & 0.6030(7) & 0.8310(1) \\ \hline
slashdot & 0.5027(5) & 0.5163(4) & 0.5619(2) & 0.5612(3) & 0.5021(6) & 0.0733(7) & 0.5673(1) \\ \hline
enron    & 0.5119(6) & 0.5299(2) & 0.5200(5) & 0.5852(1) & 0.5224(3) & 0.4259(7) & 0.5220(4) \\ \hline
ohsumed  & 0.4529(6) & 0.4546(5) & 0.4758(3) & 0.5238(1) & 0.4550(4) & 0.0910(7) & 0.4942(2) \\ \hline
tmc2007  & 0.6662(2) & 0.6703(1) & 0.6552(5) & 0.6635(3) & 0.6596(4) & 0.5561(7) & 0.6013(6) \\ \hline
bibtex   & 0.3966(5) & 0.3929(6) & 0.4130(2) & 0.4055(3) & 0.4023(4) & 0.1601(7) & 0.4372(1) \\ \hline
av. ranks & 5.3 & 4.0 & 2.7 & 2.1 & 4.1 & 6.7 & 3.1 \\ \hline\hline
\multicolumn{1}{|l|}{Significance} & \multicolumn{7}{l|}{\small\vtop{\hbox{\strut $NLDD > \{BR, SMBR, RAKEL, MLKNN\}; BR > \{MLKNN\}; SMBR > \{BR, MLKNN\};$}\hbox{\strut $ECC > \{BR, SMBR, RAKEL, MLKNN\}; RAKEL > \{BR, MLKNN\}; CBM > \{MLKNN\}$}}} \\ \hline
\end{tabular}
}
\caption{$F$-$measure$ (higher is better) averaged over 10 cross validations (with ranks in parentheses). The results from the Wilcoxon test on whether or not any two results are  statistically  significant from one another  are summarized at the bottom of the table.}
\label{tab:F}
\end{table}

\begin{table}[h]
\centering
\resizebox{\textwidth}{!}{
\begin{tabular}{|C|C|C|C|C|C|C|C|}
\hline
Data     & $BR$ & $SMBR$ & \boldmath{$NLDD$} & $ECC$ & $RAKEL$ & $MLKNN$ & $CBM$ \\ \hline
emotions & 19 & 19 & 27 & 40 & 21 & 4 & 23 \\ \hline
scene    & 37 & 38 & 88 & 104 & 57 & 112 & 195 \\ \hline
yeast    & 59 & 61 & 96 & 141 & 90 & 59 & 530 \\ \hline
medical  & 43 & 44 & 101 & 312 & 73 & 93 & 1809 \\ \hline
slashdot & 52 & 57 & 428 & 280 & 104 & 1023 & 2540 \\ \hline
enron    & 126 & 127 & 248 & 572 & 265 & 201 & 16232 \\ \hline
ohsumed  & 22834 & 22987 & 12152 & 15799 & 37872 & 10641 & 7588 \\ \hline
tmc2007  & 21376 & 22145 & 16253 & 10023 & 23252 & 27394 & 38912 \\ \hline
bibtex   & 2337 & 2466 & 2762 & 3574 & 5017 & 6280 & 48834 \\ \hline
\end{tabular}
}
\caption{Running times (seconds) on benchmark multi-label data sets}
\label{tab:time2}
\end{table}

We next look at the performance of $NLDD$ by whether or not the true labelsets were observed in the training data.
A labelset has been observed if the exact labelset can be found in the training data and unobserved otherwise.
Since $NLDD$ makes a prediction by choosing a training labelset, a predicted labelset can only be partially correct on an unobserved labelset.
Table \ref{tab:bibtex} compares the evaluation results of $BR$ and $NLDD$ on two separate subsets of the test set of the $bibtex$ data.
The bibtex data were chosen because the data set contains by far the largest percentage of unobserved labelsets (33\%) among the data sets investigated.
The test data set was split into subsets $A$ and $B$; if the labelset of a test instance was an observed labelset, the instance was assigned to $A$; otherwise the instance was assigned to $B$. For all of the four metrics, $NLDD$ outperformed $BR$ even though 33\% of the labelsets in the test data were unobserved labelsets.

\begin{table}[h]
\centering
\begin{tabular}{|l|l|l|l|l|l|l|}
\hline
\multirow{2}{*}{} & \multicolumn{2}{l|}{Subset $A$} & \multicolumn{2}{l|}{Subset $B$} & \multicolumn{2}{l|}{Total ($A \cup B$)} \\ \cline{2-7}
                  & $BR$     & $NLDD$   & $BR$     & $NLDD$   & $BR$     & $NLDD$   \\ \hline
$Hamming \ loss$  & 0.0113 & 0.0091 & 0.0250 & 0.0224 & 0.0158 & 0.0134 \\ \hline
$0/1 \ loss$      & 0.7804 & 0.7163 & 0.9958 & 1.0000 & 0.8504 & 0.8084 \\ \hline
$Multi$-$label$ $accuracy$        & 0.3807 & 0.4273 & 0.2118 & 0.1870 & 0.3259 & 0.3492 \\ \hline
$F$-$measure$       & 0.4402 & 0.4785 & 0.3065 & 0.3058 & 0.3966 & 0.4130 \\ \hline
\end{tabular}
\caption{Evaluation results on the bibtex data set by whether or not the labelset was observed (Subset $A$) or unobserved (Subset $B$) in the training data.
Subset $A$ contains 67\% of the test instances and subset $B$ contains 33\%. For $Hamming \ loss$ and $0/1 \ loss$, lower is better. For $Multi$-$label$ $accuracy$ and $F$-$measure$, higher is better.}
\label{tab:bibtex}
\end{table}

We next look at the three regression parameters the proposed method ($NLDD$) estimated (equation~\ref{eq:binomial}) for each data set in more detail.
Table \ref{tab:coef} displays the $MLE$ of the parameters of the binomial model in each data set. In all data sets, the estimates of $\beta_{1}$ and $\beta_{2}$ were all positive.
The positive slopes imply that the expected loss (or, equivalently the probability of misclassification for each label) decreases as $D_{\vec{x}}$ or $D_{\vec{y}}$ decreases.

From the values of $\hat{\beta}_{0}$ we may infer how low the expected loss is when either $D_{\vec{x}}$ or $D_{\vec{y}}$ is 0. For example, $\hat{\beta}_{0} = -3.5023$ in the $scene$ data set. If $D_{\vec{x}} =0$ and $D_{\vec{y}} = 0$, $\hat{p} = 0.0292$ because $\log{\frac{\hat{p}}{1-\hat{p}}} = -3.5023$. Hence $\hat{E}(loss) = L\hat{p} = 6 \cdot 0.0292 = 0.1752$. This is the expected number of mismatched labels for choosing a training labelset whose distances to the new instance are zero in both feature and label spaces.
The results suggest the expected loss would be very small when classifying a new instance that had a duplicate in the training data ($D_{\vec{x}} =0$) and whose labels are predicted with probability $1$ and the predicted labelset was observed in the training data ($D_{\vec{y}} = 0$).

\begin{table}[h]
\centering
\begin{tabular}{|c|c|c|c|}
\hline
Data     & $\hat{\beta}_{0}$ & $\hat{\beta}_{1}$ & $\hat{\beta}_{2}$ \\ \hline
emotions & -2.6353 & 0.0321 & 1.0912 \\ \hline
scene    & -3.5023 & 0.0134 & 1.8269 \\ \hline
yeast    & -3.9053 & 0.1409 & 0.8546 \\ \hline
medical  & -5.5296 & 0.1089 & 1.6933 \\ \hline
slashdot & -4.2503 & 0.1204 & 1.3925 \\ \hline
enron    & -3.8827 & 0.0316 & 0.7755 \\ \hline
bibtex   & -4.8436 & 0.0093 & 0.7264 \\ \hline
ohsumed  & -3.1341 & 0.0022 & 0.9855 \\ \hline
tmc2007  & -3.6862 & 0.0370 & 1.1056 \\ \hline
\end{tabular}
\caption{The maximum likelihood estimates of the parameters of equation \ref{eq:binomial} averaged over 10 cross validations}
\label{tab:coef}
\end{table}

\section{Scaling up $NLDD$}

As seen in Section \ref{NLDD}, the time complexity of $NLDD$ is dependent on the size of the training data ($N$). In particular, the term $O(N^{2}(d+L))$ makes the complexity of $NLDD$ quadratic in $N$.
For larger data sets the running time could be reduced by running the algorithm on a fraction of the $N$ instances, but performance may be affected. This is investigated next.

\begin{figure}[h]
\centering
  \includegraphics[scale=0.5]{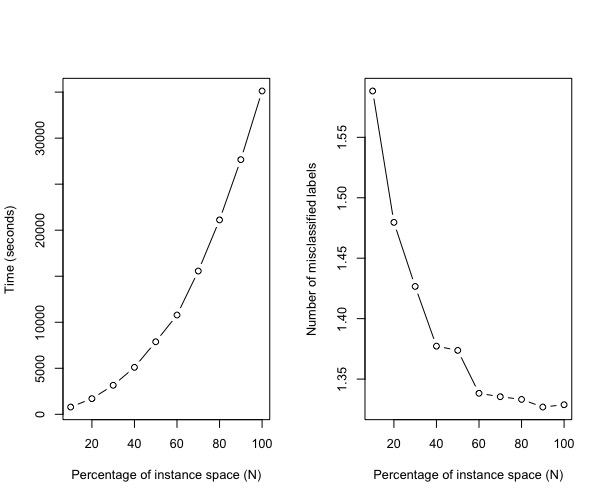}
  \caption{Running time (left) and the average number of mismatched labels (right) as a function of the percentage of the instance space for $NLDD$}
  \label{complexity}
\end{figure}

Figure \ref{complexity} illustrates the running time and the corresponding performance of $NLDD$ as a function of the percentage of $N$.
For the result, we used the $tmc2007$ data with 75/25 train/test splits.
After splitting, we randomly chose 10\% - 100\% of the training data and ran $NLDD$ with the reduced data.
As before, we used $SVM$ with a linear kernel as the base classifier.

The result shows that $NLDD$ can obtain similar predictive performances for considerably less time.
The running time increased quadratically as a function of $N$ while the improvement of the performance of $NLDD$ appeared to converge.
Using 60\% of the training data, $NLDD$ achieved almost the same performance in the number of mismatched labels as using the full training data.
Similar results were obtained on other large data sets.

\section{Discussion} \label{discussion}

For the sample data sets selected, $NLDD$ performed significantly better than $BR$, $SMBR$ and $MLKNN$ on all of the four metrics.
$NLDD$ also significantly outperformed $ECC$ on $Hamming \ loss$ and $0/1 \ loss$, $RAKEL$ on $0/1 \ loss$, $multi$-$label$ $accuracy$ and $F$-$measure$.
Although no significant difference was found between $NLDD$ and $CBM$, $NLDD$ achieved higher average ranks on all of the four metrics.
On any evaluation metric, no method performed statistically significantly better than $NLDD$.

Like $BR$, \ $NLDD$ uses outputs of independent binary classifiers.
Using the distances in the feature and label spaces in binomial regression, $NLDD$ can make more accurate predictions than $BR$.
$NLDD$ was also significantly superior to $SMBR$, which is similar to $NLDD$ in the sense that it makes predictions by choosing training labelsets using binary classifiers.
$SMBR$ is based on the label space only, while $NLDD$ uses the distances in the feature space as well.

Like $LP$, the proposed method treats each training labelset as a different class of a single-label problem in the prediction stage. Using a training labelset as a predicted vector, the proposed approach takes potentially high order label correlations into account.

In fitting the binomial regression, $NLDD$ restricts the fit of the binomial model to distance pairs with low distances in the feature and label spaces.
This dramatically reduces the size of the data used for regression fitting.
In the $yeast$ data set, the training data $T$ contained  2178 instances.
Since we equally divided the training data into $T_{1}$ and $T_{2}$, each of them contained 1089 instances.
Hence the number of possible instances available for fitting is $1089*1089 = 1,185,921$.
On the other hand, $NLDD$ used only $2,018$ instances which is less than $0.2\%$ of all instances.

$NLDD$ requires more time than $BR$.
The relative differences of running time between $NLDD$ and $BR$ depended on the size of the training data ($N$).
The number of labels and features had less impact on the differences, as the complexity of $NLDD$ is linear in them.
For the larger data sets, we reduced the running time of $NLDD$ by using a subset (70\%) of the training data.
The results of $ohsumed$ and $tmc2007$ data sets show that $NLDD$ with reduced data can perform fast compared to not only $BR$ but also the other methods on large data problems.

Because $NLDD$ makes a prediction by choosing a training labelset, the prediction label vector is confined to a labelset appearing in the training data.
If a new instance has a true labelset unobserved in the training data, there will be at least one incorrect predicted label.
Even so, $NLDD$ beat the other methods on average.
How frequently an unobserved labelset occurs depends on the data set.
For most data sets, less than 5\% of the test data contained labelsets not observed in the training data.
In other words, most of the labelsets of the test instances could be found in the training data.
However, for the bibtex data set about 33\% of the test data contained unobserved labelsets.
As seen in Table \ref{tab:bibtex}, when the true labelsets of the test instances were not observed in the training data (subset $B$), $BR$ performed slightly better than $NLDD$ in terms of $0/1 \ loss$, $multi$-$label$ $accuracy$ and $F$-$measure$.
On the other hand, when the true labelsets of the test instances were observed in the training data (subset $A$), $NLDD$ outperformed $BR$ on all of the metrics.
Combined, $NLDD$ achieved higher performances than $BR$ on the entire test data. 

$NLDD$ uses binomial regression to estimate the parameters. This setup assumes that the instances in $S$ are independent.
While it turned out that this assumption worked well in practice, dependencies may arise between the two pairs of a given $S_{i}$.
If required this dependency could be modeled using, for example, generalized estimating equations \citep{liang86}.
We examined $GEE$ using an exchangeable correlation structure. The estimates were almost the same and the prediction results were unchanged. The analogous results are not shown.

For prediction, the minimization in \eqref{eq:weight} only requires the estimates of the coefficients $\beta_{1}$ and $\beta_{2}$ which determine the tradeoff between $D_{\vec{x}}$ and $D_{\vec{y}}$.  The estimate of $\beta_{0}$ is not needed. However, estimating $\beta_{0}$ allows estimating the probability of a misclassification of a label for an instance, $\hat{\theta}$.   Such an assessment of uncertainty of the prediction can be useful. For example,  one might only want to classify instances where the probability of misclassification is below a certain threshold value.

$NLDD$ uses a linear model for binomial regression specified in \ref{eq:binomial}. To investigate how the performance of $NLDD$ changes in nonlinear models,
we also considered a model: $\log\left(\frac{\theta}{1-\theta}\right) = \beta_{0} + D_{\vec{x}}^{\beta_{1}} \cdot D_{\vec{y}}^{\beta_{2}}$ in which the distances are combined in a multiplicative way.
The difference of prediction results obtained by the linear and multiplicative models was small. The analogous results are not shown.

While $SVM$ was employed as the base classifier, other algorithms could be chosen provided
the classifier can estimate posterior probabilities rather than just scores.
Better predictions of binary classifiers will make distances in the label space more useful and hence lead to a better performance.

\section{Conclusion} \label{conclusion}

In this paper, we have presented $NLDD$ based on probabilistic binary classifiers.
The proposed method chooses a training labelset with the minimum expected loss, where the expected loss is a function of two variables: the distances in feature and label spaces. The parameters are estimated by maximum likelihood.
The experimental study with 9 different multi-label data sets showed that $NLDD$ outperformed other state-of-the-art methods on average in terms of $Hamming \ loss$, $0/1 \ loss$, $multi$-$label$ $accuracy$ and $F$-$measure$.

$NLDD$ relies on labelsets observed in the training data and is unable to predict previously unobserved labelsets. $NLDD$ outperformed other methods on the data sets observed where most test data sets contained 5\% unobserved labelsets. While the method still outperforms the other methods with 33\% of unobserved labelsets on the bibtex data, the method might not fare as well when the percentage of unobserved labelsets are substantially greater.

\section*{Acknowledgement}

This research was supported in part by National Science and Engineering Research Council of Canada (NSERC, Gweon) and by Social Sciences and Humanities Research Council of Canada (SSHRC  \#  435-2013-0128, Schonlau).

\bibliography{multilabel_ref}
\bibliographystyle{spbasic}

\end{document}